\documentclass[nonacm,sigconf]{acmart}

\usepackage{xurl}
\usepackage{nicefrac}
\usepackage{xargs}
\usepackage[capitalize]{cleveref}
\usepackage{enumitem}
\usepackage{multirow}
\usepackage[frozencache,cachedir=.]{minted}  %
\usepackage{tikz}
\usepackage{etoolbox}
\usepackage{hyperref}
\usepackage{url}

\newcommand{\sten}{STen}
\newcommand{\nofm}{\texorpdfstring{$n$:$m$}{n:m}}
\newcommand{\nofmg}{\texorpdfstring{$n$:$m$:$g$}{n:m:g}}
\newcommand{\bertbase}{BERT\textsubscript{\tiny\textsc{BASE}}}

\newcommand{\bigo}[1]{\mathcal{O}{\left({#1}\right)}}
\DeclareMathOperator*{\argmax}{arg\,max}

\definecolor{CircleBlue}{RGB}{219,232,250}
\newcommand*{\circlenum}[1]{%
  \tikz[baseline=(char.base)]\node[anchor=south west, draw=CircleBlue, circle, inner sep=0pt, fill=CircleBlue](char){\color{black}\sffamily\footnotesize {#1}};}

\newcommand{\timesh}{{\mkern-1mu\times\mkern-2mu}}

\definecolor{LightGray}{gray}{0.95}
\definecolor{STenPyColor}{RGB}{47,124,32}
\newcommand{\STenPy}{\textcolor{STenPyColor}{sten}}
\setminted{fontsize=\scriptsize, bgcolor=LightGray, escapeinside=||}
\newmintinline{python}{python3, fontsize=\footnotesize, bgcolor=LightGray}
\makeatletter
\patchcmd{\minted@colorbg}{\medskip}{}{}{}
\patchcmd{\endminted@colorbg}{\medskip}{\vspace{-0.2em}}{}{}
\makeatletter

\crefname{section}{\S}{\S\S}
\crefformat{section}{#2\S{#1}#3}
\crefrangeformat{section}{#3\S\S{#1}#4 to~#5{#2}#6}
\crefmultiformat{section}{#2\S\S{#1}#3}{ and~#2{#1}#3}{, #2{#1}#3}{ and~#2{#1}#3}
\crefname{appendix}{\S}{\S\S}
\crefformat{appendix}{#2\S{#1}#3}

\setlist[itemize]{leftmargin=*,nosep,topsep=0pt,partopsep=0pt}
\setlist[enumerate]{leftmargin=*,nosep,topsep=0pt,partopsep=0pt}
\setlist[description]{leftmargin=0.5em,nosep,labelsep=\fontdimen2\font,topsep=0pt,partopsep=0pt}
\setlist{nolistsep}

\begin{document}

\title{\sten: Productive and Efficient Sparsity in PyTorch}

\author{Andrei ivanov}
\affiliation{%
  \department{Department of Computer Science}
  \institution{ETH Z\"{u}rich}
  \country{Switzerland}
}

\author{Nikoli Dryden}
\affiliation{%
  \department{Department of Computer Science}
  \institution{ETH Z\"{u}rich}
  \country{Switzerland}
}

\author{Tal Ben-Nun}
\affiliation{%
  \department{Department of Computer Science}
  \institution{ETH Z\"{u}rich}
  \country{Switzerland}
}

\author{Saleh Ashkboos}
\affiliation{%
  \department{Department of Computer Science}
  \institution{ETH Z\"{u}rich}
  \country{Switzerland}
}

\author{Torsten Hoefler}
\affiliation{%
  \department{Department of Computer Science}
  \institution{ETH Z\"{u}rich}
  \country{Switzerland}
}

\begin{abstract}
As deep learning models grow, sparsity is becoming an increasingly critical component of deep neural networks, enabling improved performance and reduced storage.
However, existing frameworks offer poor support for sparsity.
Specialized sparsity engines focus exclusively on sparse inference, while general frameworks primarily focus on sparse tensors in classical formats and neglect the broader sparsification pipeline necessary for using sparse models, especially during training.
Further, existing frameworks are not easily extensible: adding a new sparse tensor format or operator is challenging and time-consuming.
To address this, we propose \sten{}, a sparsity programming model and interface for PyTorch, which incorporates sparsity layouts,
operators, and sparsifiers, in an efficient, customizable, and extensible framework that supports virtually all sparsification methods.
We demonstrate this by developing a
high-performance grouped \nofm{} sparsity layout for CPU inference at moderate sparsity.
\sten{} brings high performance and ease of use to the ML community, making sparsity easily accessible.

\end{abstract}


\maketitle

\section{Introduction}
\label{sec:intro}
\begin{figure*}[t]
  \centering
  \includegraphics[width=0.95\textwidth]{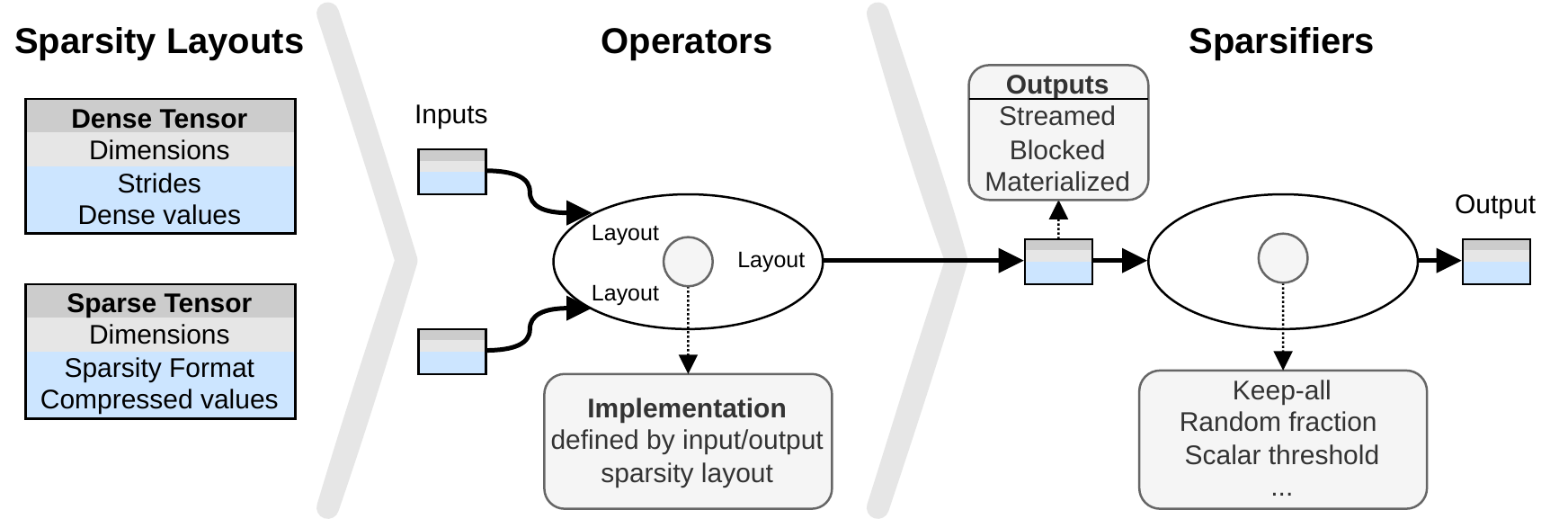}
  \caption{Overview of the \sten{} sparsity programming model.}\label{fig:overview}
\end{figure*}

Deep learning models are growing voraciously, and require ever growing amounts of compute and memory~\cite{openaicompute,sevilla2022compute,villalobos2022machine}, as larger models trained for longer continue to offer improvements~\cite{kaplan2020scaling,hoffmann2022training}.
Indeed, the largest models (e.g.,~\cite{brown2020language,chowdhery2022palm,pope2022efficiently}) do not fit in a single GPU for inference, much less training.
To address this, sparsity has emerged as a major research and engineering direction~\cite{hoefler2021sparsity,gale2019state}.
Sparsity is widely used to reduce storage requirements and improve performance during inference.
More recently, there has been interest in sparsity during training, including both sparse fine-tuning for large models and fully sparse training from scratch.

Many frameworks have begun to integrate support for sparsity.
PyTorch's~\cite{paszke2019pytorch} \texttt{torch.sparse} includes COO and CSR tensors and a limited set of operations; TensorFlow's~\cite{tensorflow2015-whitepaper} \texttt{tf.sparse} similarly supports COO tensors.
SparseML and DeepSparse~\cite{kurtz2020inducing} provide recipes and implementations for sparsity algorithms, and a highly-tuned sparse inference engine for CPUs, respectively.
Other frameworks, such as TVM~\cite{octomlsparse} and TFLite~\cite{tflitesparse}, also support sparse inference.

However, sparsity is typically ``tacked on'' to general deep learning frameworks: existing frameworks offer only limited support for sparsity, and lack a complete and flexible sparsification pipeline.
Frameworks like TensorFlow and PyTorch primarily support sparse tensors, and do not offer general support for sparse operations (e.g., \texttt{torch.sparse} supports only 13 of 44 linear algebra operations, of which four support backpropagation, and hundreds of other operations are also not supported).
They also lack native support for sparsification operations that can efficiently produce sparse tensors (e.g., without materializing intermediates).

Further, at sparsities common in deep learning (50--95\%), although the supported classical sparse matrix formats reduce storage (COO, CSR), they often perform worse than dense implementations.
It is also challenging to add support for additional formats, especially if automatic differentiation is to be supported.
While blocked formats (e.g., ELL, BCSR) support efficient implementations by calling dense kernels for each block~\cite{gale2020sparse}, they restrict where nonzeros can be placed and can limit the information preserved after sparsification.
Such implementations (e.g.,~\cite{openaiblocksparse,pytorchblocksparse}) also typically support only a handful of customized operations.
Finally, many implementations use masks to emulate sparsity by zeroing out elements; while valuable for research, this offers no storage or performance improvements and the masking is typically implemented by hand.

Thus, existing frameworks offer limited productivity and efficiency improvements for sparsity, and do not provide a clear path toward supporting a broad use of sparsity.
This is especially acute for researchers working at the forefront of sparsity, who want to rapidly iterate on novel sparsity formats, fast implementations, and sparsification methods, yet are stymied by frameworks.

To address this, we first propose a new programming model for sparsity in PyTorch (\cref{sec:sten}), overviewed in \cref{fig:overview}.
It consists of three components: \emph{sparsity layouts} for tensors; \emph{operators}, which provide implementations for computations with any combination of sparsity layouts for input and output tensors; and \emph{sparsifiers}, which are applied to operator outputs to compute a new sparse tensor.
Sparsifiers are further classified as \emph{streaming}, \emph{blocking}, or \emph{materializing}, based on the number and structure of outputs they require.
Our model supports the vast majority of sparsification approaches, and enables them to be implemented efficiently; for example, threshold pruning is a streaming sparsifier, and a high-performance implementation could be inlined into operators.

We provide an implementation of this model, \textbf{\sten{}}\footnote{https://github.com/spcl/sten}, in PyTorch (\cref{sec:impl}).
\sten{} provides a comprehensive sparsification pipeline and supports fast sparse inference as well as sparse GPU training using masking, for all PyTorch operators.
\sten{} is fully extensible, and a user or developer can easily implement additional sparsity layouts, operators, or sparsifiers.
Backpropagation is transparently supported.
If there is no implementation for a sparse operator or sparsity layout, \sten{} automatically converts to a supported implementation to ease development.
This allows \sten{} to provide a productive and efficient playground for sparsity in PyTorch, allowing researchers to pursue novel ideas without significant framework engineering overhead.
To demonstrate this flexibility, we develop a novel \emph{grouped} \nofm{} sparsity layout (\nofmg{}, \cref{sec:groupednofm}), building on existing \nofm{} formats~\cite{ampere,zhou2021learning}, for fast sparse inference at moderate sparsity levels.
We also include a suite of standard sparsifiers.

\sten{} delivers competitive performance for both sparse inference and training without sacrificing its flexibility (\cref{sec:eval}).
We evaluate our \nofmg{} sparsity layout for sparse inference against state-of-the-art inference engines, where we are up to 3.9$\timesh$ faster than DeepSparse~\cite{kurtz2020inducing} using unstructured sparsity for sparse-dense GEMM.
We also show end-to-end performance for \bertbase{}~\cite{devlin2019bert}, where sparse \nofmg{} models recover the original accuracy and are 3.2$\timesh$ faster than a dense PyTorch version.
Finally, we showcase the productivity \sten{} brings.
It supports torchvision's~\cite{torchvision} entire suite of classification models and common HuggingFace transformer models~\cite{wolf2020transformers} out-of-the-box, and we illustrate how it is simple to sparsify existing models, run sparse fine-tuning, and integrate additional sparsity layouts.
\sten{} brings both high performance and ease of use to the machine learning community, enabling users to easily leverage sparsity, engineers to develop optimized sparse formats and operators, and researchers to quickly explore new sparsification methods.

\section{Sparsification Background}
\label{sec:sparsity}
Sparsity occurs when some values in a tensor are zero.
When many values are zero, it can be more efficient to only store the nonzero values, saving space.
It can also be more efficient to skip computations involving zeros, as the output is already known.
Hence, exploiting sparsity, if done correctly, can reduce memory and compute requirements.
It can also reduce data movement, a particular bottleneck in deep learning~\cite{ivanov2021data}.
The \textit{sparsity} of a tensor is defined as the ratio of the number of zeros to the size of the tensor; the number of nonzeros is referred to as $nnz$.
In general, for sparsity to pay off in performance, a tensor must be sufficiently sparse~\cite{han2017ese,yu2017scalpel}; however, the sparsity necessary to achieve improvements will vary depending on the workload, sparsity structure, hardware architecture, and implementation.\\
\includegraphics[trim={0 0 0 1.75cm},clip]{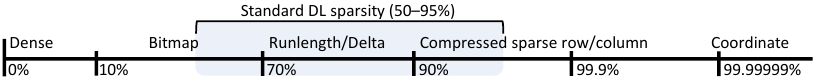}\\
We focus on sparsity in deep learning, where sparsities in the range of 50--95\% are common (see \cite{hoefler2021sparsity} for a comprehensive overview).

Sparsity typically occurs in DL workflows in three regimes:
\vspace{-3\parskip}\begin{enumerate}
\item Sparse inference, where a network is already sparse.\\
  \includegraphics[trim={0 0 1.4cm 1.9cm},clip,height=1.5em]{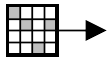}\vspace{-0.3em}
\item Sparse fine-tuning, where sparsity is induced in a pretrained dense network through a retraining process.\\
  \includegraphics[trim={0 0 0.8cm 1.9cm},clip,height=1.5em]{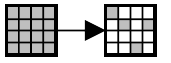}\vspace{-0.3em}
\item Sparse training, where a sparse network is trained from scratch, possibly with the sparsity changing over time.\\
  \includegraphics[trim={0 0 0 1.9cm},clip,height=1.5em]{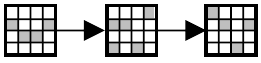}\vspace{-0.3em}
\end{enumerate}\vspace{-\parskip}
We distinguish between fine-tuning, which is typically a relatively short process on a pretrained network, and training from scratch, typically a much longer process.

There are many works that aim to accelerate sparse inference in particular (e.g.,~\cite{kurtz2020inducing,octomlsparse,ye2022sparsetir}).
Such sparse models are usually constructed by using sparsifiers to prune the weights of an existing dense model in a sparse fine-tuning stage.
Both sparse fine-tuning and training typically use dense tensors in combination with masking to emulate sparsity, especially as the sparsity pattern may change during training.
Fully sparse training remains an open problem, although there has been recent progress on this, including iterative pruning~\cite{zhu2017prune,frankle2018lottery} and training from scratch (e.g.,~\cite{evci2019difficulty,dao2022monarch,evci2022gradmax}).

Beyond sparsity in weights, other tensors in a network may also be sparse.
This includes activations (e.g., through dropout~\cite{srivastava2014dropout}, ReLUs~\cite{nair2010rectified}, or other methods~\cite{kurtz2020inducing,hunter2022two}) and gradients (common in communication compression, e.g.,~\cite{strom2015scalable,dryden2016communication,lin2017deep}).

Sparsity is induced using sparsifiers, which select the values to prune (i.e., set to zero).
A simple, yet powerful, class of sparsifiers which we consider is magnitude pruning~\cite{zhu2017prune,frankle2018lottery}, which prunes small values on the assumption that they will not change the output significantly and so are unimportant.
There are several varieties of magnitude pruning.
One-shot magnitude pruning prunes a network to a desired sparsity in a single step, after which the network may be fine-tuned.
Iterative magnitude pruning alternates between pruning and fine-tuning stages while gradually increasing the sparsity level.
The pruning may also be local or global; the former prunes each layer to the desired sparsity, while the latter considers the entire network.
This can allow the pruning to better allocate parameters throughout the network.
Another variant is layer-wise magnitude pruning, which prunes layers one at a time in sequential order, fine-tuning after each.

Numerous formats exist for efficiently representing sparsity, although the level and structure of the sparsity influences which formats yield speedups~\cite{pooch1973survey}.
Frameworks like PyTorch support formats such as Compressed Sparse Row (CSR), which represents nonzeros using a matrix of column and row indices; and Coordinate Offset (COO), which stores nonzeros together with their absolute offset.
Approaches like ELLPACK (ELL) and block CSR (BCSR) more efficiently store blocked data.
Recently, specialized formats for DL workloads have been developed, such as \nofm{}~\cite{ampere,zhou2021learning}, where each group of $m$ elements has $n$ nonzeros (e.g., $2$:$4$ \raisebox{-0.1em}{\includegraphics[trim={0 0 0 1.9cm},clip,height=\fontcharht\font`\B]{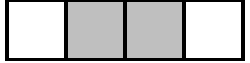}}).

\section{The \sten{} Programming Model}
\label{sec:sten}
We now introduce our programming model for sparsity; see \cref{fig:overview} for an overview.
The core components of the model are \emph{sparsity layouts} (\cref{subsec:model-layouts}), \emph{operators} (\cref{subsec:model-ops}), and \emph{sparsifiers} (\cref{subsec:model-sparsifiers}), which together can represent sparse operations and sparsification.
Here we describe the high-level concepts and interface, and discuss implementation details in \cref{sec:impl}.
Although our focus is on PyTorch, we expect the \sten{} model to be portable to other frameworks.
The key aim of this model is to support a complete and performant sparsity pipeline with extensive customizability for researchers, both those working on sparsity and on performance.

\subsection{Sparsity Layouts}\label{subsec:model-layouts}

Sparsity layouts augment the typical memory layout of a tensor by annotating the sparsity format used to store the data.
The user can specify any sparsity layout, including classic formats like CSR, CSC, and COO; blocked formats like ELL and BCSR; specialized formats like \nofm{}; or dense tensor masking; plus any associated parameters (e.g., $n$ and $m$).
A sparsity layout is assigned to a tensor as follows:
\begin{minted}{python3}
x = |\STenPy{}|.torch_tensor_to_csr(sparsifier, x)
\end{minted}
Further, custom sparsity formats can easily be added as well:
\begin{minted}{python3}
class CscTensor:  # Use SciPy sparse CSC matrix
  def __init__(self, data):
    self.data = scipy.sparse.csc_matrix(data)
  def to_dense(self):
    return torch.from_numpy(self.data.todense())
\end{minted}
Programmers simply define the \texttt{to\_dense} operation and one sparsifier (see \cref{subsec:model-sparsifiers}) to enable sparse/dense conversion for the format.
If in-place operations, which directly modify an existing tensor, are to be supported, an additional implementation needs to be defined for \sten{}'s \texttt{SameFormatSparsifier} to handle sparsification (see \cref{sec:impl}).

This contrasts with existing frameworks, where adding support for new sparsity layouts is a complex and time-consuming process that often involves significant additions to the framework ``core''.
\sten{}'s flexibility enables researchers to rapidly prototype and evaluate novel sparsity layouts and performance engineers to easily tune the layouts of individual tensors for maximum performance.

\subsection{Operators}\label{subsec:model-ops}

Operators are any function (e.g., matrix multiplication), with any number of input and output tensors (we ignore non-tensor arguments for simplicity).
We allow the input and output tensors to have any sparsity layout, and an operator may have different implementations for each combination of tensor layouts for maximum performance.
Calling a built-in operator is simple:
\begin{minted}{python3}
a = |\STenPy{}|.torch_tensor_to_csr(sparsifier, torch.randn(4, 4))  # Sparse
b = torch.randn(4, 4)  # Dense
c = torch.mm(a, b)  # Dispatched to sparse(CSR)*dense->dense mm implementation
\end{minted}
To produce sparse output, an operator must be associated with a sparsifier.
We discuss such sparse operators in \cref{subsec:model-sparsifiers}.
Beyond this, there are no restrictions or additional requirements for operators.

As it can be infeasible to provide implementations for every combination, several approaches to simplify this are provided (see \cref{subsec:impl-ops}).
A pattern matching system for supported input/output sparsity layout combinations can dispatch to a single underlying implementation.
Otherwise, the interface falls back to a dense implementation with masks and issues a warning.
This enables incremental optimization by performance engineers, who can decide which operators it is most profitable to provide sparse implementations for.
It also allows users to rapidly explore the effects of sparsity on models (albeit with a performance penalty).

\subsection{Sparsifiers}\label{subsec:model-sparsifiers}

\begin{table}[t]
  \centering%
  \caption{Sparsifier types and examples, number of passes mode over a tensor, memory requirements ($nnz$ total nonzeros, block size $b$ when blocking), and sparsifier type.
    Some complex weight sparsifiers could be implemented more efficiently than with materialization.}\label{tab:sparsifiers}%
  \footnotesize%
  \begin{tabular}{@{}l@{\hspace{\tabcolsep}}l@{\hspace{0pt}}r@{\hspace{\tabcolsep}}l@{\hspace{\tabcolsep}}r@{}}%
    \toprule
    Sparsifier & Examples & Passes & Memory & Type \\
    \midrule
    Keep-all & Sparse add & 1 & $\bigo{1}$ & Streaming \\
    Random fraction & Dropout~\cite{srivastava2014dropout} & 1 & $\bigo{1}$ & Streaming \\
    Scalar threshold & ReLU~\cite{nair2010rectified} & 1 & $\bigo{1}$ & Streaming \\
    Per-block fraction & \nofm{}~\cite{ampere,zhou2021learning} & 2 & $\bigo{b}$ & Blocking \\
    Scalar fraction & Magnitude~\cite{zhu2017prune,frankle2018lottery} & 2 & $\bigo{nnz}$ & Materializing \\
    Block-wise fraction & Block magnitude~\cite{li2016pruning} & 2 & $\bigo{nnz}$ & Materializing \\
    \multirow{2}{*}{Complex weight sparsifiers} & Movement~\cite{sanh2020movement}, $\ell_0$~\cite{louizos2017bayesian}, & \multirow{2}{*}{$\geq$ 1} & \multirow{2}{*}{$\bigo{nnz}$} & \multirow{2}{*}{Materializing} \\
    & and others~\cite{hoefler2021sparsity} & & & \\
    \bottomrule
  \end{tabular}%
\end{table}

Sparsifiers decide which output values to keep.
Each output tensor of an operator is associated with a sparsifier.
A sparsifier can be thought of as a special kind of operator, and may include additional inputs, which may delay its application until they are ready (e.g., gradients for first-order sparsification).
Sparsifiers may also produce output in a different sparsity layout than what the associated operator outputs.
We use the term \emph{sparse operator} to refer to the combination of an operator and sparsifier.

We classify sparsifiers as one of \emph{streaming}, \emph{blocking}, or \emph{materializing}, depending on the amount of data they need before they can begin to produce output.
\vspace{-\parskip}\begin{description}
\item[Streaming] sparsifiers decide whether to prune each output value in a single pass before writing to the output tensor.\\
  \includegraphics[trim={0 0 0 1.9cm},clip,height=1.5em]{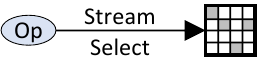}\vspace{-0.3em}
\item[Blocking] sparsifiers require a small set of output values to decide which ones to drop.\\
  \includegraphics[trim={0 0 0 1.9cm},clip,height=1.5em]{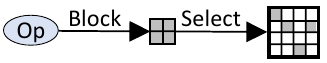}\vspace{-0.3em}
\item[Materializing] sparsifiers require the operator to fully store all values before pruning can be performed.\\
  \includegraphics[trim={0 0 0 1.9cm},clip,height=1.5em]{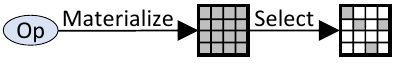}\vspace{-0.3em}
\end{description}\vspace{-\parskip}
Different types of sparsifiers support different optimizations.
E.g., streaming sparsifiers could be fused into their associated operator; this may also be possible with blockin sparsifiers.
Sparsifiers may also be used standalone to convert dense tensors to sparse.

\Cref{tab:sparsifiers} lists example sparsifiers and their characteristics.
The trivial \emph{keep-all} sparsifier preserves all produced values and is the default for dense tensors.
It is not limited to dense tensors, however: the sum of two sparse tensors with a keep-all sparsifier produces a new sparse tensor with nonzero values given by the union of the nonzeros of the inputs.
A \emph{random fraction} sparsifier drops values with a fixed probability, while a \emph{scalar threshold} sparsifier drops them if they are less than a fixed threshold.
\emph{Per-block fraction} drops the smallest proportion of values within fixed blocks of elements.
\emph{Scalar fraction} drops the smallest portion of the values (i.e., magnitude pruning) and \emph{block-wise fraction} drops entire blocks with the smallest combined absolute magnitude.
Finally, more advanced \emph{complex weight sparsifiers} require additional information such as the loss or gradients.
All of these sparsifiers could be supported by \sten{}.
Further, these examples are not exhaustive, and \sten{} can support nearly any sparsifier.

Declaring a new sparsifier is straightforward in \sten{}:
\begin{minted}{python3}
class RandomFractionSparsifier:
  def __init__(self, fraction):
    self.fraction = fraction

@|\STenPy{}|.register_sparsifier_implementation(
  sparsifier=RandomFractionSparsifier,
  inp=torch.Tensor, out=CscTensor)
def dense_to_csc_random_fraction(sparsifier, tensor, grad_fmt=None):
  # Implementation for dense->CSC sparsification
\end{minted}
We now discuss sparse operators, which associate a sparsifier with an operator to enable it to produce sparse output.
For maximum performance and flexibility, a sparse operator requires an \emph{output format}, which consists of an inline sparsifier, a temporary sparsity layout, an external sparsifier, and the output sparsity layout.\\
\includegraphics[trim={0 0 0 1.2cm},clip]{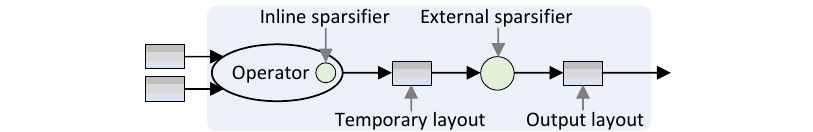}\\
The inline sparsifier is inlined into the operator and produces output in the temporary layout, which is materialized before being sparsified by the external sparsifier, which produces the final output.
Streaming or blocking sparsifiers are good candidates for implementation as inline sparsifiers, to benefit from inlining, whereas materializing sparsifiers are typically implemented as external sparsifiers.
One typically does not need both an inline and external sparsifier; however, highly optimized implementations of two-pass sparsifiers may use an inline sparsifier for the first pass and an external sparsifier for the second pass.
Sparse gradients are similarly specified for operators.
Defining a sparse operator in \sten{} simply requires specifying the original operator and output format:
\begin{minted}{python3}
# Output is CSC, sparsified w/ RandomFractionSparsifier
sparse_add = |\STenPy{}|.sparsified_op(
  orig_op=torch.add,
  out_fmt=[
    (|\STenPy{}|.KeepAll(), torch.Tensor,
     RandomFractionSparsifier(0.5), CscTensor)],
  grad_out_fmt=tuple([...])) # Similar to out_fmt

# Call the operator
c = sparse_add(a, b)
\end{minted}

\subsection{Constructing Sparse Models}\label{subsec:model-cons}

With the core components of our sparsity programming model in place, we now describe the process for building sparse models.
We consider two cases: constructing a model from scratch and sparsifying an existing model.
In short, a sparse model is set up by providing a list of tensors and the desired sparsity layout for each.
Given this, \sten{} can initialize tensors and dispatch operators to appropriate implementations.
If an operator implementation is not available, the user can provide one, convert tensors to a supported sparsity layout, or fall back to a dense implementation.

In this model, all tensors are used as operator inputs, outputs, or both; for simplicity we will ignore sparsity for model inputs and outputs.
Tensors that are input-only during forward propagation are typically \emph{weights}.
All other tensors are used as both inputs and outputs and occur within the computation graph (e.g., activations); we refer to these as \emph{intermediate tensors}.
In practice in most frameworks (including PyTorch), intermediate tensors do not exist until runtime, so their sparsity layout is instead defined by the operator that produces them.

\textbf{Constructing a sparse model} from scratch is similar to the typical process in PyTorch, but tensors and operators in its computational graph are annotated with specific sparsity layouts and, if needed, sparsifiers.
For example, one can construct a simple sparse linear layer as follows:
\begin{minted}{python3}
class SparseLinear(torch.nn.Module):
  def __init__(self, in_f, out_f, wsparsity):
    super().__init__()
    self.w = |\STenPy{}|.SparseParameterWrapper(
      dense_to_csc_random_fraction(
        RandomFractionSparsifier(wsparsity),
        torch.randn(in_f, out_f),
        # Output format for gradients:
        (|\STenPy{}|.KeepAll(), torch.Tensor,
         RandomFractionSparsifier(wsparsity),
         CscTensor)))

  def forward(self, x):
    return torch.matmul(x, self.w) # Calls sparse impl
\end{minted}

\textbf{Sparsifying an existing model} requires marking a subset of the model's tensors as sparse.
While this is straightforward for weights, sparsifying intermediate tensors is more challenging.
Unlike when building a model from scratch, we cannot mark operators as sparse: this would require modifying or rewriting the original model definition, a significant overhead for the user.
Doing so may also break compatibility with existing saved model checkpoints, complicating sparse fine-tuning.
To address this, \sten{} supports tracing an existing model to identify its intermediate tensors and operators (see \cref{subsec:impl-create}), which can then be marked with the desired sparsity layouts.
As an example:
\begin{minted}{python3}
# Sparsify a parameter:
sb = |\STenPy{}|.SparsityBuilder()
sb.set_weight(
  'net.weight',  # Traced name
  initial_sparsifier=RandomFractionSparsifer(0.9),
  out_format=CscTensor)
# Sparsify an intermediate activation:
sb.set_interm(
  'net.gelu',  # Traced name
  inline_sparsifier=RandomFractionSparsifier(0.9),
  tmp_format=CscTensor,
  external_sparsifier=KeepAll(),
  out_format=CscTensor)
# Create sparse version of original model:
snet = sb.get_sparse_model(net)
\end{minted}

To sparsify existing dense weights, or load sparse weights, we need only the desired sparsity layout and sparsifier.
We first trivially convert sparsifiers to a materializing version if needed.
Then the weights are sparsified and subsequent operator calls will use the sparse version.
For iterative sparsification methods, the sparsity can be adjusted as sparse fine-tuning proceeds.
Intermediate tensors are sparsified at runtime, as they do not exist in advance.

For training, error signals (gradients of the loss w.r.t.\ layer output) and gradients may also be sparse, and can have independent sparse layouts and sparsifiers from their associated forward pass tensors.
These are marked using a similar API: \texttt{sb.set\_weight\_grad} and \texttt{sb.set\_interm\_grad}.
We treat these identically to intermediate tensors in the forward pass.
Note that during training, weight tensors are no longer input-only, as gradient updates are applied to them.
This is not a significant change from the user perspective, and mainly implies that materializing sparsifiers may be less efficient for weights and that sparsifying on-the-fly with the gradient update operator may be faster.

\section{\sten{} Implementation}
\label{sec:impl}
\begin{figure}[t]
  \centering
  \includegraphics[width=0.95\linewidth]{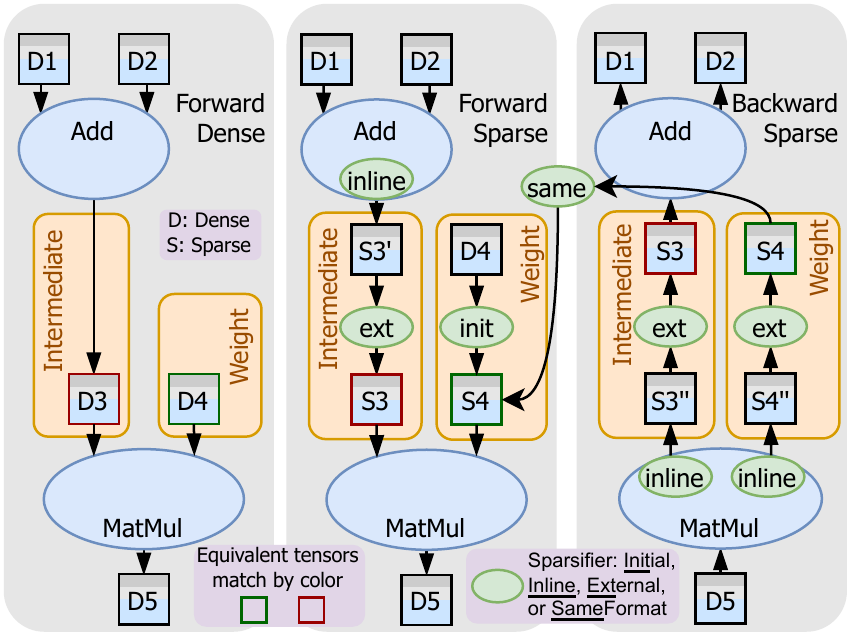}
  \caption{Example sparsification of a model with \sten{}, originally an Add followed by a MatMul (left).
    \sten{} constructs sparse forward (center) and backward (right) passes.}\label{fig:sten_interm_weight}
\end{figure}

\begin{figure*}[t]
  \centering
  \includegraphics[width=0.95\textwidth]{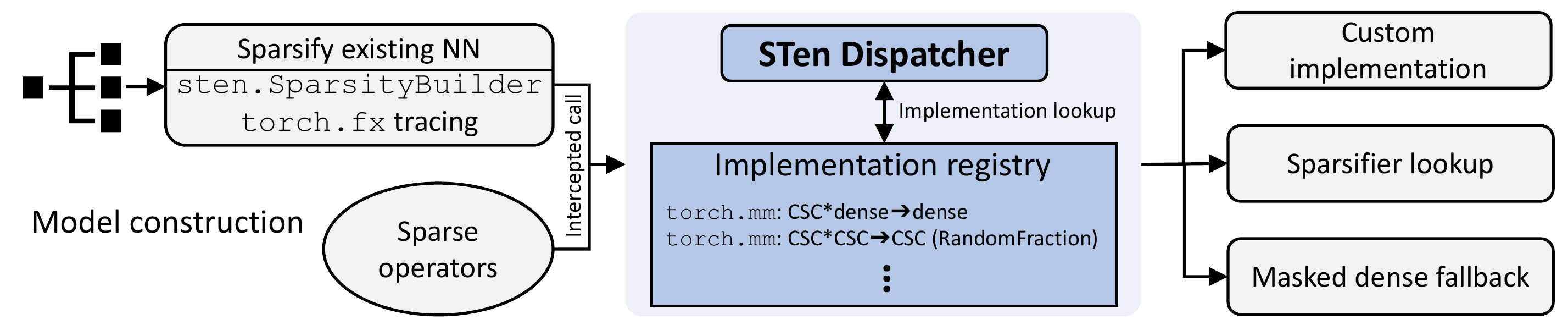}
  \caption{Overview of \sten{}'s sparse dispatch engine.}\label{fig:impl-overview}
\end{figure*}

We now discuss the implementation of \sten{} in PyTorch.
\sten{} builds upon and extends existing PyTorch infrastructure to support sparsity in a familiar manner.

\Cref{fig:sten_interm_weight} illustrates how \sten{} modifies the computation graph to support the interface specified in \cref{sec:sten}.
Depending on the location in the graph of the tensor(s) to be sparsified (i.e., intermediate or weight), the semantics of the forward pass differ.
Intermediate tensors require both inline and external sparsifiers, as detailed in \cref{subsec:model-sparsifiers}.
Weight tensors only require an initialization sparsifier.

During the forward pass, the Add operator incorporates sparsifiers.
However, during the backward pass, sparsifiers need to be integrated into the MatMul operator to provide sparsified gradients for the weights.
Given that the tensor and its gradient may have different formats, the operator that normally performs in-place modifications of the weights must now calculate the updated weights into a new tensor.
The new tensor is sparsified using the \texttt{SameFormatSparsifier} to maintain the same format it had before.

\subsection{Creating an \sten{} Model}\label{subsec:impl-create}

There are two mechanisms for creating an \sten{} model: constructing a model from scratch using sparse operators, or replacing weights or operators in an existing model with sparse versions (\cref{subsec:model-cons}).
In either case, models still follow the standard \texttt{torch.nn.Module} style.
However, operations are automatically dispatched to sparse implementations by \sten{} when sparse input tensors or operators are used (\cref{subsec:impl-ops}), so code can be written using standard operators.

Incorporating sparse weights into a model is straightforward, as PyTorch \texttt{Parameter}s are easily accessible and modifiable.
Replacing operators with sparse versions or sparsifying intermediate tensors is more challenging, as operators may not exist as standalone objects (e.g., \texttt{torch.nn.functional} methods), and intermediate tensors only exist at runtime.
We use the \texttt{torch.fx} symbolic tracer to obtain the complete computation graph of the network, and then replace operators with \sten{} wrappers that use our operator dispatcher.
Unlike weights, intermediate tensors are sparsified by associating a sparsifier with the operator that outputs them.

\subsection{Sparse Tensors}\label{subsec:impl-tensors}

Adding new sparse tensor formats to PyTorch is challenging, especially given that we want \sten{} to support arbitrary sparsity layouts.
While PyTorch supports custom \texttt{Tensor} types~\cite{extendingpytorch}, none of the approaches are suitable for \sten{}.
The key issue is that one cannot separate tensor metadata and values in the context of PyTorch's autograd engine, the core of which is written in C++.
As it is currently implemented, the C++ component of autograd expects that gradient shapes are the same as that of the corresponding parameters and that the parameters have all their data materialized.
This does not hold with \sten{}, as the tensors may have different sparsity layouts and we do not wish to materialize them at each iteration.
PyTorch's own sparse tensors avoid this issue by being directly supported within the autograd engine.
This is also not suitable for \sten{}, as we support any user-provided format from Python; rebuilding PyTorch would be infeasible for users and modifying its core time-consuming for developers.

We work around this by wrapping all user-provided sparse tensors into a single-element dense PyTorch tensor.
This ensures PyTorch's autograd can call the correct backward operator implementations when needed, without adding overhead or requiring modifications to PyTorch.
The \sten{} interface ensures the dummy tensors behave like tensors of the correct shape, so this is entirely hidden from users and all operations function as expected.

\subsection{Sparsifiers and Sparsity Layouts}\label{subsec:impl-sparsifiers}

\sten{} imposes no restrictions on the implementation of sparsity layouts or sparsifiers, which may leverage existing libraries, custom-optimized implementations, or pure Python implementations (for research and prototyping).
The API is designed with the assumption that the user declares classes that store the desired metadata (e.g., sparsity level, block size) and then registers implementations for particular input layouts.
Registered implementations are treated as a black box for maximum flexibility.
\sten{} relies on the implementer to manage the internal structure and metadata of these objects.

For sparsity layouts in particular, dense values and location metadata (e.g., indices of nonzeros) may be stored.
Sparse operator implementations (\cref{subsec:impl-ops}) are then expected to call optimized (e.g., native C++ or CUDA) implementations and pass relevant data.
We encourage implementers to rely on PyTorch's memory management primitives (e.g., storing nonzeros in dense PyTorch tensors) to simplify code and benefit from memory management optimizations.

\subsection{Sparse Operators and Dispatch}\label{subsec:impl-ops}

\begin{figure}[t]
  \centering
  \includegraphics[width=0.95\linewidth]{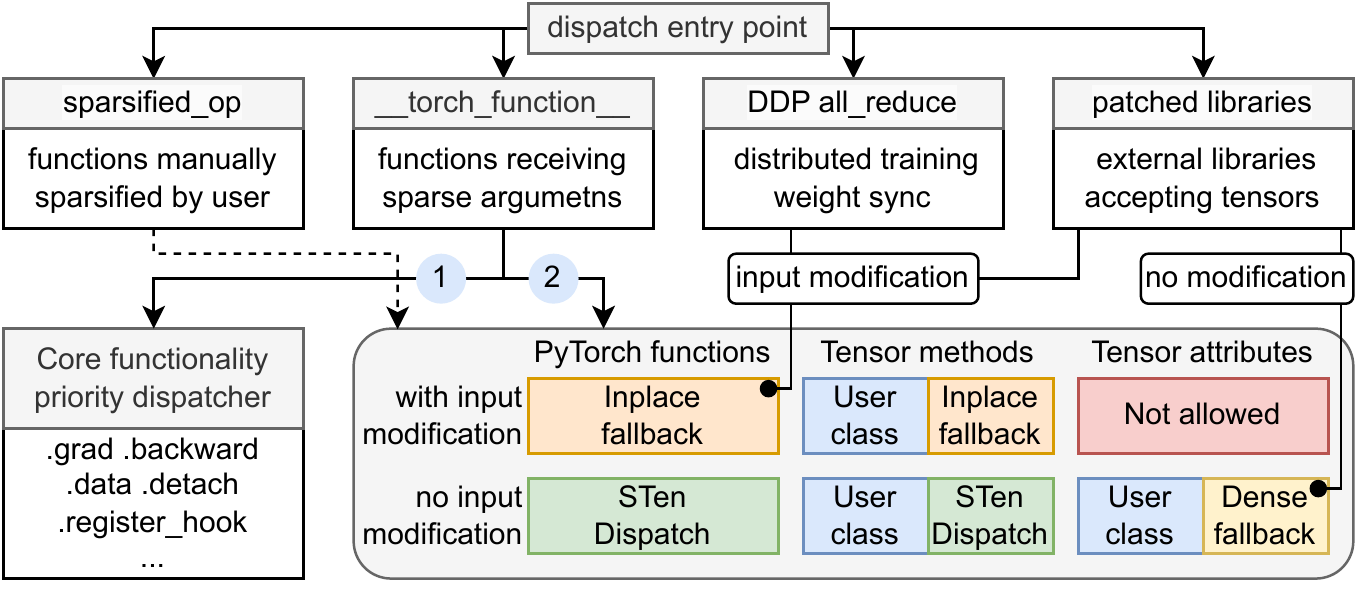}
  \caption{Overview of \sten{}'s dispatcher in PyTorch.}\label{fig:sten_dispatch_detailed}
\end{figure}

\sten{}'s operator dispatch is the key component that ties sparsity layouts, operators, and sparsifiers together.
Sparse operators may be implemented in any way desired (see \cref{subsec:impl-sparsifiers}), and the dispatch mechanism will call the appropriate implementation for the given combination of input and output sparsity layouts and sparsifiers.
Further, \sten{} will also intercept calls to standard PyTorch operations using sparse inputs and dispatch them to specialized implementations.
\cref{fig:impl-overview} provides an overview of the \sten{} dispatcher, while \cref{fig:sten_dispatch_detailed} details its integration with PyTorch.

To customize implementations for a specific sparse tensor class in \sten{}, the most straightforward way is to override the method or attribute (e.g., \texttt{shape}) by defining it with the same name as in the corresponding dense tensor.
\sten{} will search for the \emph{user class} implementation first before proceeding to other dispatch cases.

We use PyTorch's standard tensor extension mechanisms to handle calls to simple operations (e.g., \texttt{torch.add}, \texttt{torch.mm}) using sparse matrices. If sparse implementations are not found, the inputs are automatically converted to \emph{dense fallbacks}.
More complicated operations, such as in-place modifications (e.g., \texttt{add\_}) or views are handled similarly, although \sten{}'s default implementations may be pessimistic, necessitating the resparsification of the original tensor in the \emph{inplace fallback} implementation.

There is an additional global route to \sten{}'s sparse dispatcher, which is used for operators that are not supported by PyTorch's tensor extension mechanism.
This is primarily to support external libraries that provide extensions in the form of native, compiled implementations (e.g., Nvidia Apex~\cite{apex}).
While such libraries are implemented via autograd function extensions, PyTorch does not catch calls to their implementations with custom tensor types.
To support such libraries, we provide a patching API that redirects any Python function (including native built-ins) into our dispatcher when one of their arguments is a sparse tensor.
Users need only request such patching for non-standard extension libraries, as \sten{} supports common ones by default.

\sten{} does \emph{not} allow the use of tensor attributes that trigger in-place modifications on the applied tensor.
This restriction is in place to prevent potential interactions with PyTorch's implementation details, such as autograd functionality.
\sten{} manages such operations internally within its \emph{priority dispatcher for core functionality}. The priority operators are given precedence \circlenum{1} and are looked up first. The dispatcher section, which is aware of operator semantics (inplace or not) and its type (function, method, or attribute), handles all the remaining operators \circlenum{2}.

When execution reaches \sten{}'s \emph{dispatcher}, \sten{} uses information about the original operator, input and output sparsity layouts, and sparsifiers to look up the sparse implementation in a global registry.
This is done independently for forward and backward passes, allowing separate implementations to be chosen.
The lookup simply involves a hash table search based on a canonicalized list of input and output sparsity layouts and sparsifiers, a search which can easily be rapidly pruned using the layouts.
If no implementation is found, \sten{} can attempt to convert sparse tensors to alternative sparsity layouts and rerun the dispatch lookup.
Conversion is only attempted when \sten{} can guarantee that it is lossless, to prevent any information loss; generally this means it only attempts conversion to formats such as CSR.
Finally, if no other implementation is found, \sten{} will fall back to a dense implementation by converting sparse tensors to dense tensors with masks and applying standard dense operator implementations.
Note that, depending on the hardware and implementations, a dense implementation may outperform unoptimized sparse implementations, so users may prefer this to other conversions.
Hence, \sten{} automatically supports sparse implementations of \emph{all} PyTorch operators (albeit with some performance overhead when conversion is needed, see \cref{sec:eval}).

\sten{} does not implement its own sparse operators, and relies on users (or downstream libraries) to register their own implementations.
There are no restrictions on implementations, which may leverage existing libraries or custom implementations.
Support for common sparse operators (e.g., linear layers) using the implementations in \texttt{torch.sparse} and \texttt{scipy.sparse}, as well as our new \nofmg{} format, are provided by default.
Additionally, the operators are custom PyTorch operators and interface with standard PyTorch features such as JIT compilation.
We emphasize that a user is \emph{not} expected to provide optimized implementations for all combinations of sparsity layouts, sparsifiers, hardware, and the like, and indeed this is likely to be infeasible.
Rather, \sten{} enables users to implement the key operators which their particular problem requires, while \sten{}'s dispatcher will use conversion or dense implementations for any unsupported operators, allowing rapid prototyping and incremental optimization.

\subsection{Backpropagation}\label{subsec:impl-backprop}

\sten{} is fully integrated with PyTorch's autograd engine.
We leverage a custom autograd extension (via \texttt{torch.autograd.Function}), which lets PyTorch manage calling the correct backward implementation.
Custom implementations of backward operators can be provided, and \sten{} will fall back to dense versions (with conversion) if implementations are not available.
If a custom forward implementation is provided for an operator, we require that a backward pass implementation be provided as well if backpropagation is to be supported (as the default implementation may expect different preserved states than what the custom one provides).

\begin{figure}[t]
  \centering
  \includegraphics[trim={0.7cm 0.1cm 0 0},clip,width=\linewidth]{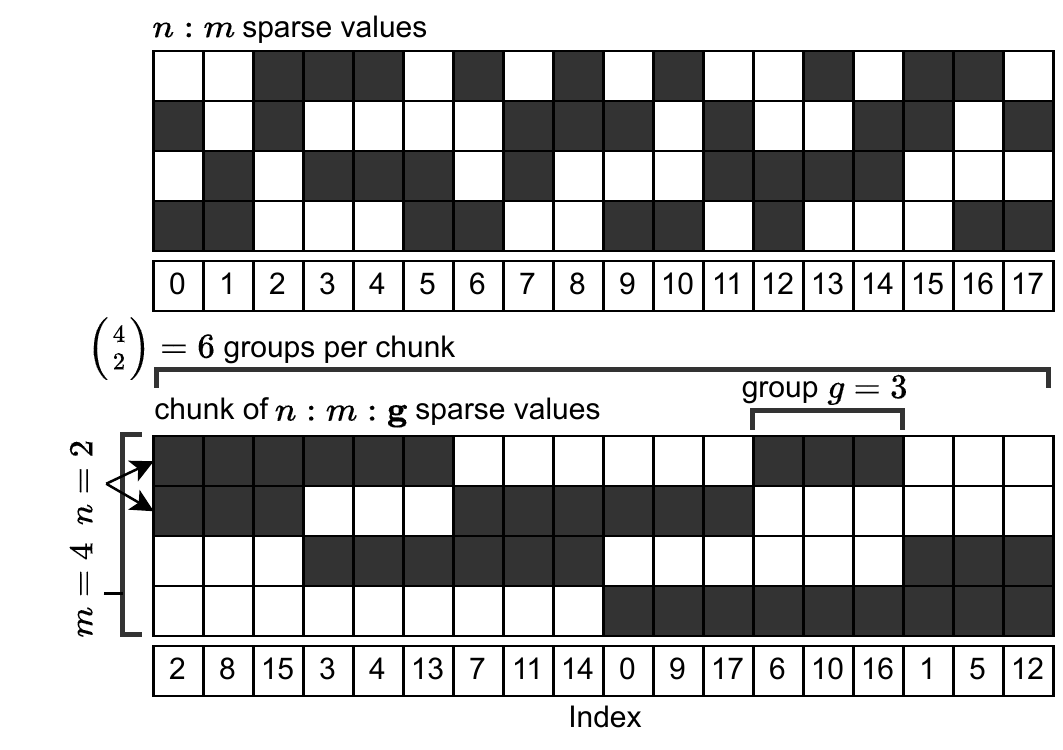}
  \caption{Example illustrating one \emph{chunk} of our new grouped \nofm{} sparsity format (\nofmg) with $n=2$, $m=4$, and $g=3$.}\label{fig:grouped_n_m}
\end{figure}

\subsection{Distributed Training}\label{subsec:impl-dist}

\sten{} natively supports distributed data-parallel training using PyTorch's \texttt{DistributedDataParallel} module.
This is handled internally by converting sparse gradient tensors to dense before utilizing native collective operations, and finally converting the tensors back.
We implement this using \sten{}'s operator patching mechanism.

As tensor conversion may be expensive for some formats, we support optimized converters.
In particular, for structured sparsity formats, we avoid unnecessary conversions when the nonzero locations of the initial and replacement tensors match.
This naturally appears during gradient synchronization, as tensors have the same nonzeros on every processor at the beginning of training, and the user can optionally ensure the sparsity pattern remains fixed or changes slowly.
These optimizations are beneficial because it can be cheaper to convert from sparse to dense formats and compare the nonzero layout than to convert from dense to sparse.

Future work in \sten{} could allow it to support fully sparse gradient communication (e.g.,~\cite{dryden2016communication,strom2015scalable,lin2017deep,renggli2019sparcml,li2022near}) or more general communication compression~\cite{tang2020communication} for additional performance.

\section{Grouped \nofm{} Sparsity (\nofmg{})}
\label{sec:groupednofm}
\begin{figure}[t]
  \centering
  \includegraphics[width=\linewidth]{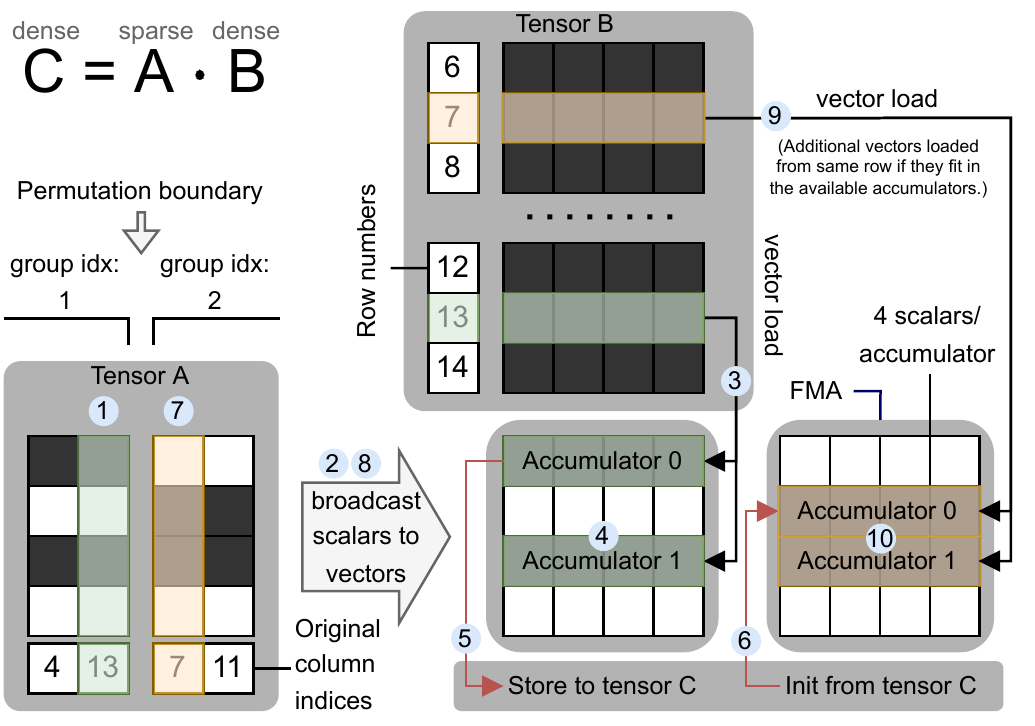}
  \caption{Sparse-dense GEMM kernel for $3$:$6$:$g$ sparsity format.}\label{fig:grouped_n_m_gemm}
\end{figure}

To illustrate the flexibility of \sten{}, we develop a novel sparsity layout, a \emph{grouped} \nofm{} format that enables fast CPU inference at moderate sparsity.
The format is an extension of existing \nofm{} sparsity formats~\cite{ampere,zhou2021learning} with the addition that each nonzero pattern is repeated $g$ times, forming a \emph{group}, for easier mapping to vector operations for performance.
Additionally, we combine groups into \emph{chunks} with all $\binom{m}{n}$ combinations of nonzeros in  fixed order.
For flexibility in the encoding, we permit reordering the blocks of $m$ elements within each chunk, and so store an index encoding the original location of each block.
This means that, even with one group, we impose more structure than in standard \nofm{}.
We refer to this as \nofmg{} sparsity, and illustrate it in \cref{fig:grouped_n_m}.
Below we discuss its implementation and performance considerations, and we evaluate accuracy and performance in \cref{sec:eval}.

\subsection{\nofmg{} Implementation}\label{subsec:nofmg-impl}

The design of the \nofmg{} sparsity layout is heavily influenced by performance considerations.
\Cref{fig:grouped_n_m_gemm} illustrates our \nofmg{} sparse-dense GEMM kernel.
We implement high-level optimizations such as caching, tiling, and parallelism following OpenBLAS~\cite{wang2013augem,xianyi2012model}, and utilize an optimized microkernel to implement the complete GEMM algorithm.
\circlenum{1} Data is first loaded from sparse values, then \circlenum{2} broadcast into vector registers.
Chunks, which fix the order of sparsity permutations, allow kernels to avoid branches based on the sparsity structure.
\circlenum{3} The corresponding dense data is then loaded using indirect loads from specific rows of $B$, \circlenum{4} and a fused multiply-add (FMA) is performed.
We provide implementations using both AVX2 and AVX-512 vector operations.
Finally, when a permutation boundary is reached, \circlenum{5} the result is stored in $C$.
\circlenum{6}--\circlenum{10} perform the next iteration.
By using fixed group sizes, we can efficiently unroll iterations over groups and map them to vector registers.
The permutation order in chunks is selected so the nonzero pattern between adjacent groups differs in only one location, so that we need save and initialize only one vector register.
Overall, this implementation delivers high performance even at moderate sparsities.

The choice of $n$ and $m$ influence the efficiency of the implementation.
In particular, as the number of permutations $\binom{m}{n}$ in a chunk grows, the implementation may be less efficient (e.g., due to limited instruction cache space).
Nevertheless, we can efficiently support sparsity between 50 and 95\%.

\subsection{Constructing \nofmg{} Sparse Tensors}\label{subsec:nofmg-cons}

We now propose algorithms for converting dense tensors to \nofmg{} sparsity layouts on both CPU and GPU.
When converting, we attempt to preserve elements to maintain the highest overall magnitude of the tensor.
More formally, for a dense tensor $X$, we wish to find $\argmax_{\hat{X}}\|\hat{X}\|$ where $\hat{X}$ is in \nofmg{} format and $\|\cdot\|$ is a matrix norm (we use $L_1$).
Performance is critical, as the primary use of these conversions is sparsifying weights after gradient updates during training.
While this could be formulated as an integer programming problem with unknowns representing the specific choice of sparsity pattern for each block, we found this to be too slow, and instead utilize fast approximations.

To sparsify on CPUs, we first compute the total magnitude of the preserved elements for each column of a chunk with all permutations of nonzero patterns.
With a fixed \nofmg{} format, there are $\binom{m}{n}^2 g$ such columnwise magnitudes.
This list is then sorted and processed from highest to lowest.
We use a specific nonzero pattern for a column in the dense tensor only if this column was not yet selected and the group corresponding to the pattern is not yet full.

To sparsify on GPUs, we use a variant of this algorithm that assigns each column of a chunk to a separate GPU thread.
Initially, columns are arbitrarily assigned to groups.
Each thread iterates over the columns in the other sparsity groups and attempts to exchange the nonzero pattern it is assigned with an alternative nonzero pattern.
If such a swap improves the overall magnitude for the pair of columns, it is performed atomically.
This continues until no changes are made.

The opposite conversion, from \nofmg{} to dense is simpler, and requires a single iteration over the values, reordering their location according to the stored index.

\subsection{Integration with \sten{}}\label{subsec:nofmg-sten}

Finally, we show key details for implementing \nofmg{} sparsity in PyTorch with \sten{}.
We first declare a new sparsity layout (\cref{subsec:model-layouts}).
\begin{minted}{python3}
class GroupedNMTensor:
  def __init__(self, val, idx, nm_strides):
    self.val, self.idx, self.nm_strides = val, idx, nm_strides
  @staticmethod
  def from_dense(tensor, n, m, sparse_dim, group_size, group_dim):
    # Call optimized dense->n:m:g conversion routine
    val, idx, nm_strides = dense_to_grouped_n_m(
      tensor, n, m, sparse_dim, group_size, group_dim)
    return GroupedNMTensor(val, idx, nm_strides)
  def to_dense(self):
    # Call optimized n:m:g->dense conversion routine
    return grouped_n_m_to_dense(self.nm_strides, self.val, self.idx)
\end{minted}
For sparse fine-tuning and training, we also make use of masked dense tensors, implemented via \texttt{FixedMaskTensor}. We next define the sparsifier and register implementations (\cref{subsec:model-sparsifiers}).
\begin{minted}{python3}
class GroupedNMSparsifier:
  def __init__(self, n, m, g):
    self.n, self.m, self.g = n, m, g
@|\STenPy{}|.register_sparsifier_implementation(
  sparsifier=GroupedNMSparsifier,
  inp=torch.Tensor, out=FixedMaskTensor)
def dense_to_grouped_n_m(sparsifier, tensor, grad_fmt=None):
  ... # Optimized sparsifier implementation
@|\STenPy{}|.register_sparsifier_implementation(
  sparsifier=|\STenPy{}|.SameFormatSparsifier,
  inp=torch.Tensor, out=FixedMaskTensor)
def same_format_grouped_n_m_and_fixed(sparsifier, tensor, grad_fmt=None):
  ... # Optimized sparsifier implementation
\end{minted}
We now define implementations for a sparse linear operator (\cref{subsec:model-ops}) using \nofmg{}.
These definitions are used for sparse training with masked dense tensors; a similar approach defines operators natively using the \nofmg{} format for sparse inference.
\begin{minted}{python3}
@|\STenPy{}|.register_fwd_op_impl(
  operator=torch.nn.functional.linear,
  inp=(torch.Tensor, FixedMaskTensor, torch.Tensor),
  out=[(|\STenPy{}|.KeepAll, torch.Tensor)])
def impl(ctx, inputs, output_sparsifiers):
  ...  # Forward implementation

@|\STenPy{}|.register_bwd_op_impl(
  operator=torch.nn.functional.linear,
  grad_out=[torch.Tensor],
  grad_inp=[(|\STenPy{}|.KeepAll, torch.Tensor),
            (|\STenPy{}|.KeepAll, FixedMaskTensor),
            (|\STenPy{}|.KeepAll, torch.Tensor)])
def impl(ctx, grad_outputs, input_grad_sparsifiers):
  ...  # Backward implementation
\end{minted}
This suffices to enable full use of \nofmg{} sparsity within PyTorch for inference and training.
Linear operators will use masked dense tensors during training, with conversions implemented with our optimized conversion routines, and our sparse-dense GEMM kernel during inference.
All other operators will use masked dense implementations through \sten{}'s dispatcher fallbacks.

\section{Evaluation}
\label{sec:eval}
\begin{figure}[t]
  \centering
  \includegraphics[width=\linewidth]{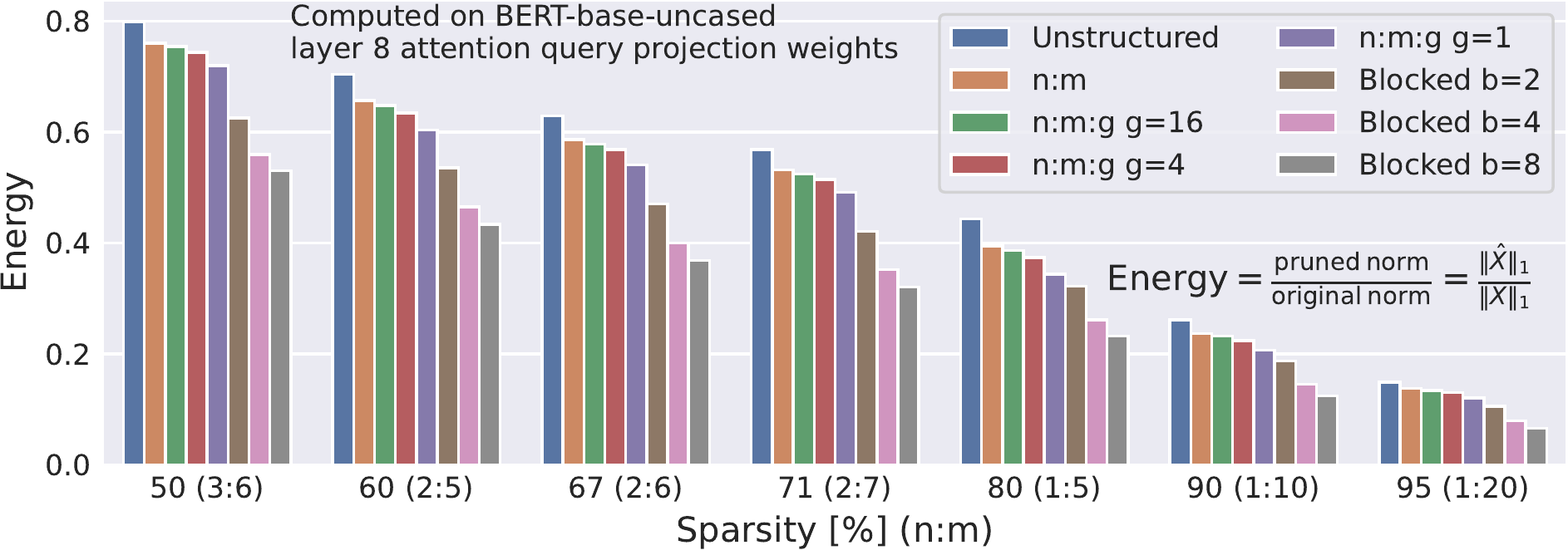}
  \caption{Trade-off between sparsity structure and accuracy for unstructured, \nofm{}, \nofmg{}, and blocked sparsity.}\label{fig:bert-energy}
\end{figure}

Here we first consider the performance and accuracy of our \nofmg{} sparsity layout, and then showcase the productivity of \sten{} by implementing multiple different sparsifiers.

All experiments used PyTorch~\cite{paszke2019pytorch} 1.12 and CUDA 11.4.
CPU implementations for \nofmg{} sparsity were compiled with GCC 8.4.1.

\subsection{\nofmg{}: Performance and Accuracy}

We now evaluate our \nofmg{} sparsity layout.
First, we study how well it preserves tensors compared to sparsity layouts with more or less structure.
We then show that it can prune \bertbase{}~\cite{devlin2019bert} with little loss in accuracy.
Finally, we evaluate the performance of our pruned BERT models for sparse inference.

\textbf{\nofmg{} structure.}
Generally, there is a trade-off in pruning: adding structure to sparsity can improve performance, but may lower accuracy, as it is more restrictive.
We therefore compare \nofmg{} with varying group size to less (unstructured and \nofm{}) and more (blocked) structured sparsity on individual tensors.
To do this, we compare the \emph{energy}, a novel metric we define to be the ratio $\|\hat{X}\|_1 / \|X\|_1$, where $X$ is the original tensor and $\hat{X}$ is the pruned version.
The energy ranges between 0 and 1, and captures the intuition that we want to preserve larger-magnitude values.

In \cref{fig:bert-energy}, we show the energy for a weight tensor from the HuggingFace~\cite{wolf2020transformers} \texttt{bert-base-uncased} model.
The trends are nearly identical for other layers, as well as layers from ResNet-50~\cite{he2016deep}.
Unstructured sparsity preserves the maximum possible energy, followed by \nofm{}.
Our \nofmg{} sparsity achieves almost the same energy as \nofm{} when $g=16$, with quality decreasing slightly with smaller $g$.
This is because increasing group size tends to be less restrictive: as chunks are larger, it is easier to find more \nofm{} patterns.
Blocked sparsity performs the worst.
Hence, we can see that \nofmg{} sparsity preserves energy nearly as well as \nofm{} sparsity, while its additional structure can be taken advantage of for performance.

\begin{figure}[t]
  \centering
  \includegraphics[width=\linewidth]{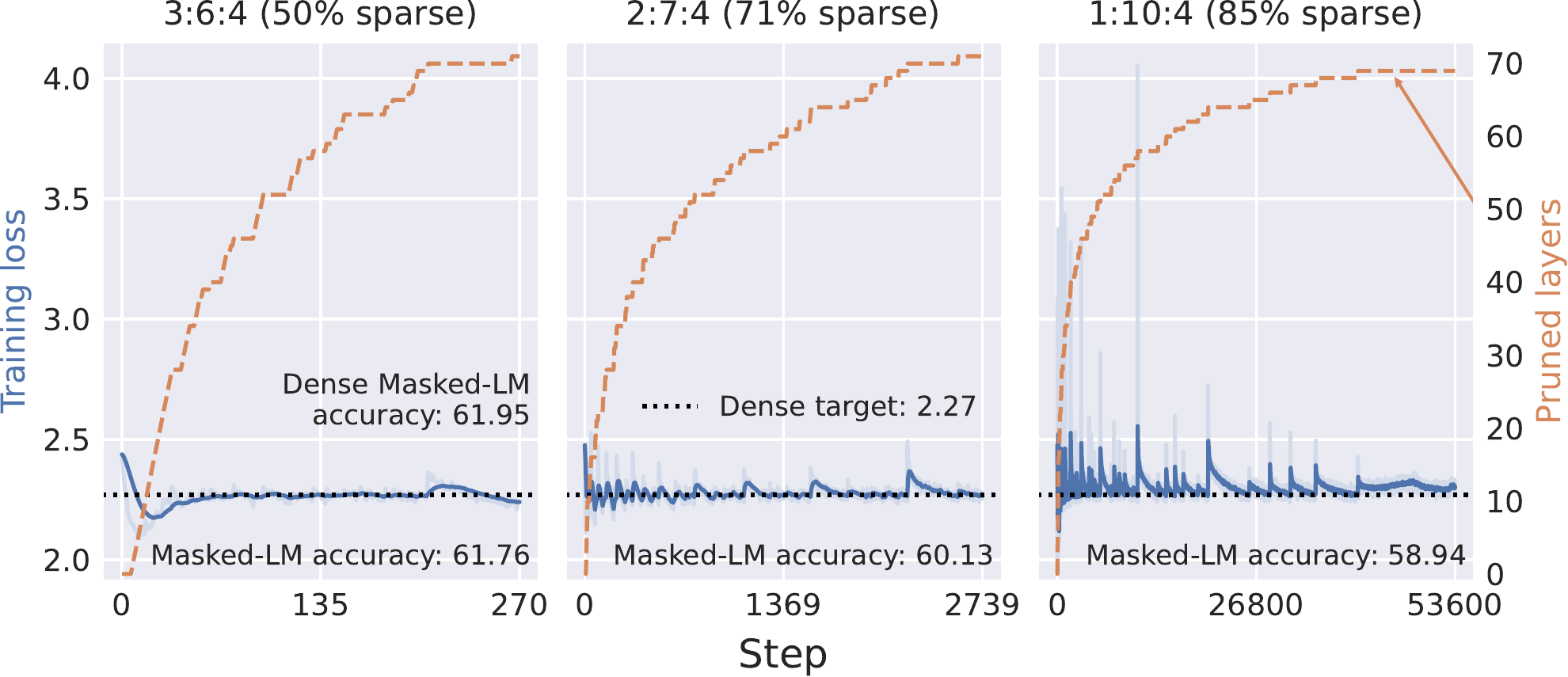}
  \caption{\bertbase{} training loss with \nofmg{} sparse pruning.}
  \label{fig:step_layers_loss}
\end{figure}

\textbf{Sparse fine-tuning.}
We now evaluate the quality of \nofmg{} sparsity by sparsifying a dense, pretrained \bertbase{} model (HuggingFace \texttt{bert-base-uncased}) using iterative layer-wise magnitude pruning.
Each layer is sparsified, and then the model is fine-tuned following standard BERT pretraining until the model recovers its original loss.
This process continues until either all 72 layers are sparsified or the loss ceases to diminish. In the latter case, the remaining weights are left dense, making the overall sparsity smaller than the per-layer target.
We use the Wikipedia and BookCorpus datasets, a constant learning rate of $5\cdot 10^{-4}$, and a global effective batch size of 4096 with gradient accumulation.
We use three RTX3090 GPUs (local batch size 16); when all layers are sparse, each step takes a median of 42 s (compared to 18 s when dense).

\Cref{fig:step_layers_loss} shows the training loss over the course of pruning and reports the final test masked language modeling accuracy. During training using the 1:10:4 format, we reduced the loss to $10^{-4}$ after step 48,000, where the spike in the loss curve occurred.
Our sparse models are able to maintain good performance, although fine-tuning for greater sparsity requires more training iterations.

\begin{figure}[t]
  \centering
  \includegraphics[width=\linewidth]{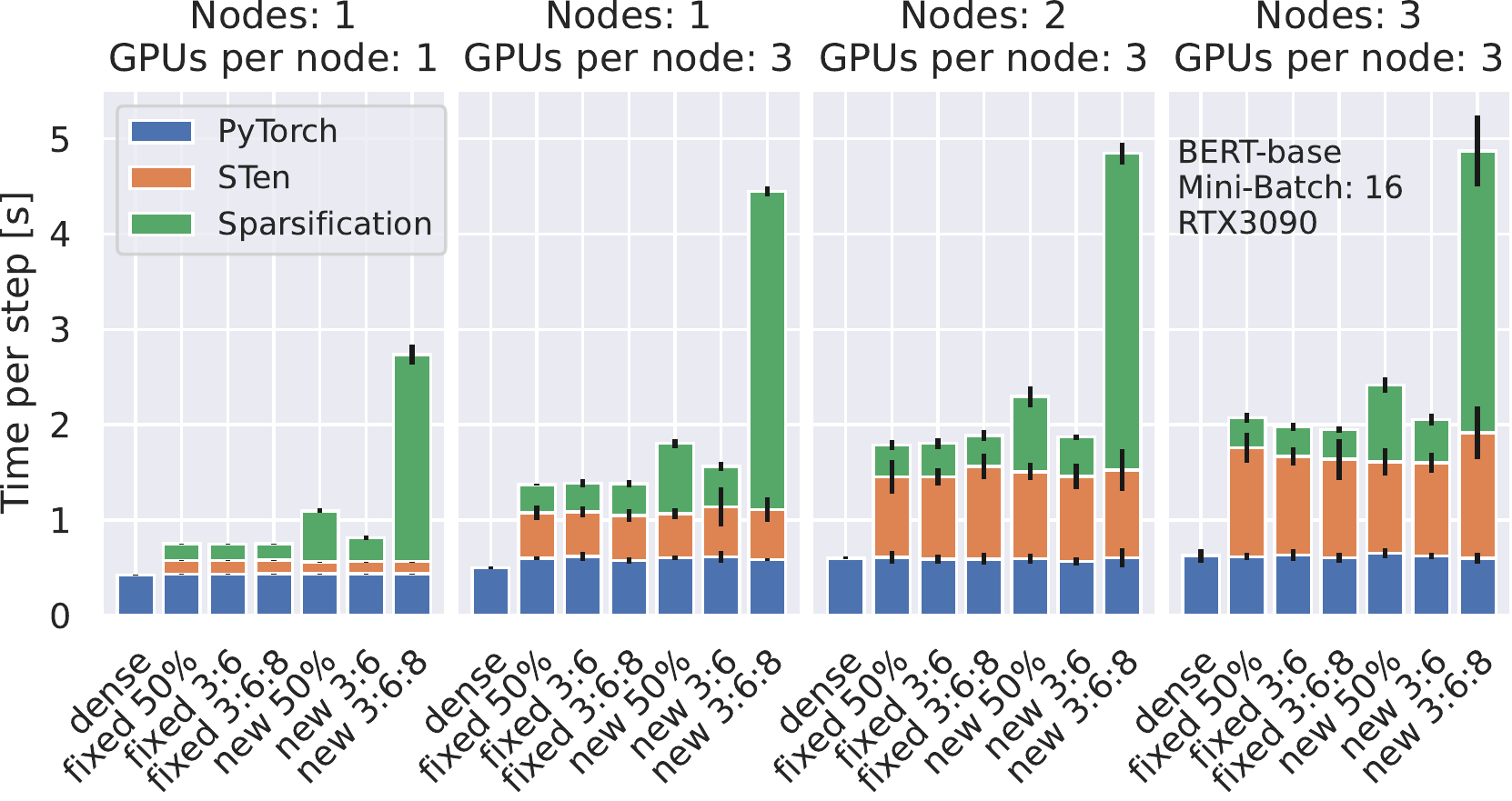}
  \caption{Masked training overheads for unstructured, \nofm{}, and \nofmg{} formats. \emph{Fixed} sparsification keeps the same non-zero mask, while \emph{new} sparsification recomputes the mask.}
  \label{fig:overhead_analysis}
\end{figure}

In \cref{fig:overhead_analysis}, we show the overheads added by masked sparse training with \sten{}.
During training, GPU kernels are faster than the CPU kernels used for inference, so \sten{} dispatch overheads are more noticeable, although further optimization could mitigate this.
As \nofmg{} is optimized for CPU inference and we make use of masking during training, there is necessarily some overhead compared to dense training.
In most iterations during training, the sparsity mask is \emph{fixed} as the sparsity pattern changes slowly, which reduces sparsification overhead.
Recomputing the mask (e.g., when sparsity increases) is more expensive for formats with complex constraints.
Distributed training is another source of overheads as tensors on different model copies may have different masks due to independent sparsification.
To synchronize weights, collective operations on sparse tensors need to be performed, which we implement by converting tensors to dense, exchanging their values, and rerunning sparsification with an optimized conversion path (see \cref{subsec:impl-dist}).

We conducted an evaluation of the overheads associated with distributed sparse training by performing weak scaling across varying numbers of GPUs, while keeping the mini-batch size fixed at 8. The experiments were conducted on 128 nodes of the Piz Daint supercomputer, utilizing the Intel Xeon E5-2690v3 CPU (2.60GHz), and one NVIDIA Tesla P100 GPU with 16GB of memory per node. 

Scaling efficiency, computed as the ratio of training runtime on 1 and 128 GPUs, went from 40\% (0.52 s on 1 GPU vs 1.32 s on 128 GPUs) for dense training to 30\% (1.02 s on 1 GPU vs 3.37 s on 128 GPUs) for masked sparse training. Despite the conservative handling of sparse tensors by requiring conversion to dense and resparsification on every step, STen exhibits less than 10\% of weak scaling overhead in total.

\begin{figure}[t]
  \centering
  \includegraphics[width=\linewidth]{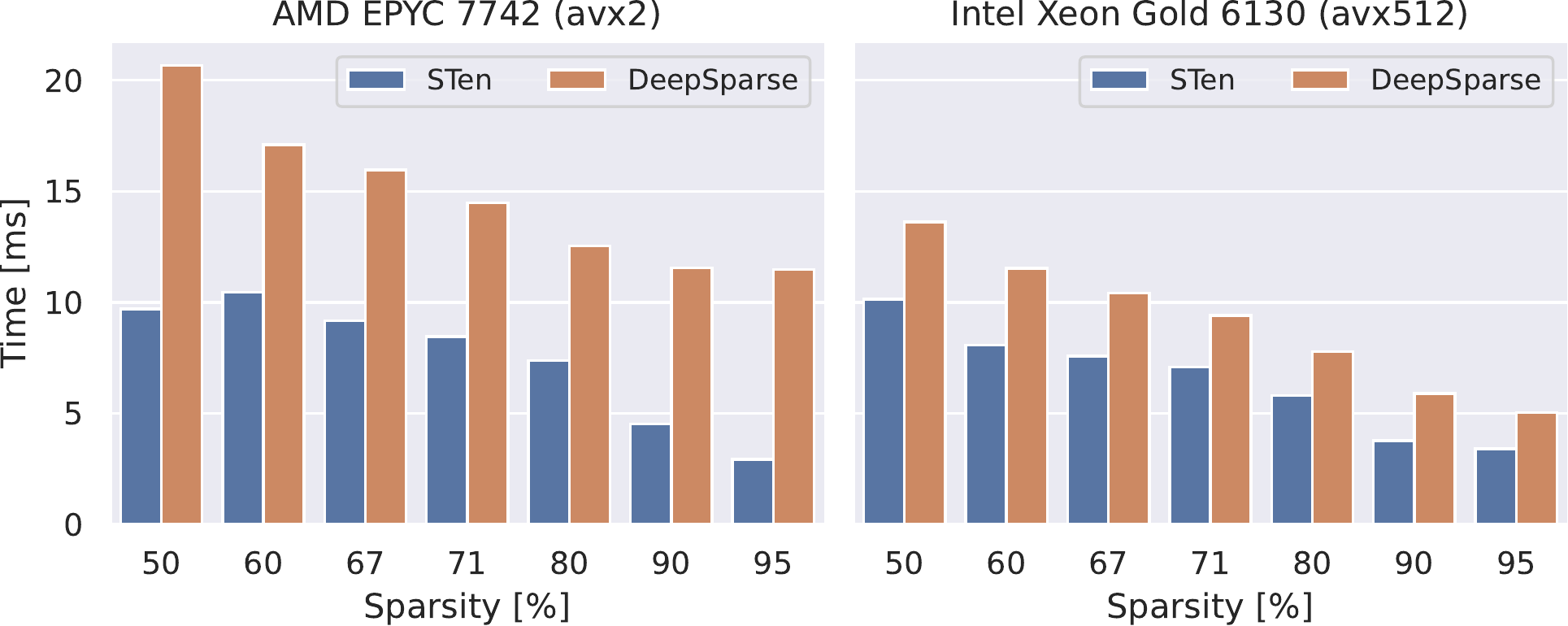}
  \caption{Median runtime of our \nofmg{} sparse-dense GEMM compared with the DeepSparse optimized inference engine.}
  \label{fig:nmg-microbench}
\end{figure}

\begin{figure}[t]
  \centering
  \includegraphics[width=\linewidth]{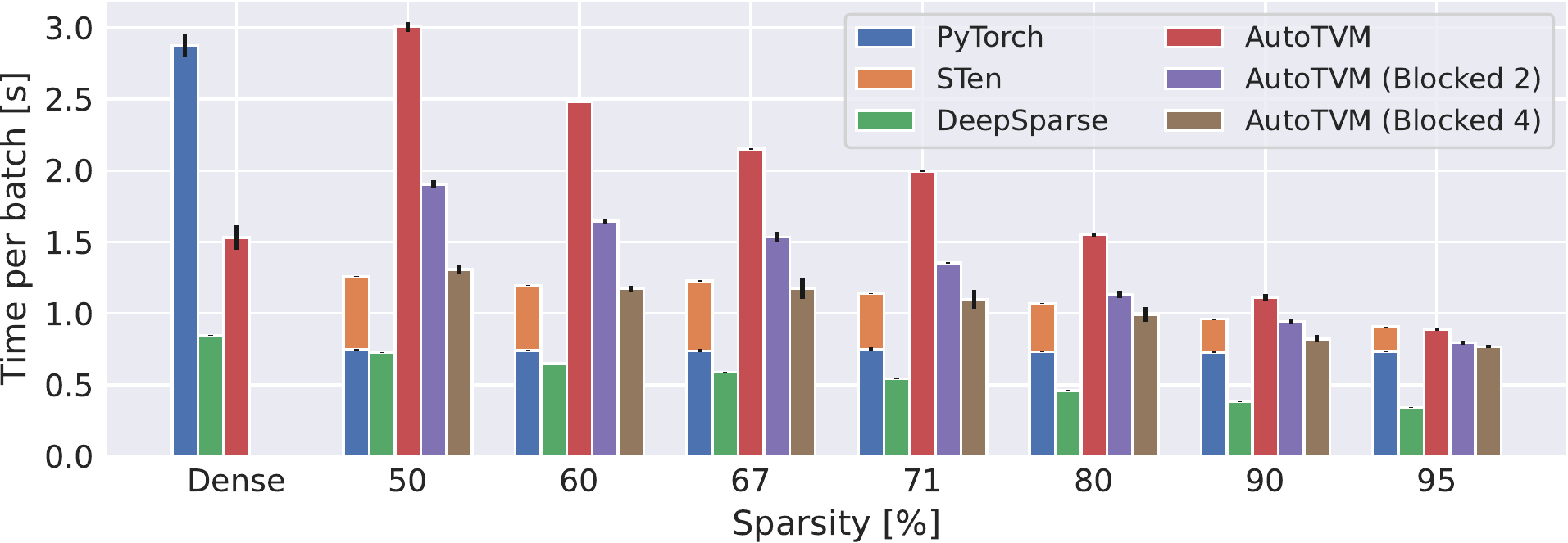}
  \caption{Median sparse \bertbase{} inference latency.}
  \label{fig:bert_inf_perf}
\end{figure}

\textbf{Sparse inference.}
We first evaluate the performance of our \nofmg{} sparse-dense GEMM (\cref{subsec:nofmg-impl}), and compare it with Deep\-Sparse \cite{kurtz2020inducing} (v1.1.0) on a sparse unstructured tensor.
\Cref{fig:nmg-microbench} shows results for a $768 \timesh 3072 \timesh 4096$ GEMM from a \bertbase{} feed-forward layer run on an AMD EPYC 7742 64-core processor (3.3 GHz) and an Intel Xeon Gold 6130 16-core processor (2.8 GHz).
Our \nofmg{} implementation is faster than DeepSparse at every sparsity level.
Overall, we see that the added structure in \nofmg{} sparsity enables significant performance gains: up to 3.9$\timesh$ on AMD and 1.5$\timesh$ on Intel.

We next show end-to-end sparse inference with \bertbase{} using the same configurations as above in \cref{fig:bert_inf_perf}.
We use batch size 8 and sequence length 512.
We also compare with TVM~\cite{chen2018tvm} (v0.10.0) using either an unstructured or block pruned tensor.
We tuned using AutoTVM~\cite{chen2018learning} but observed it did not tune sparse operators.

We see that \sten{} improves performance over the dense baseline by up to 3.2$\timesh$, with a similar speedup pattern as in the sparse-dense GEMM.
\sten{} also matches or outperforms TVM up to 90\% sparsity, but performs worse than DeepSparse or blocked TVM at higher sparsity.
This is due to framework overheads in PyTorch; \cref{fig:bert_inf_perf} breaks down our performance into \sten{} and PyTorch runtime.
While inference engines are able to eliminate these overheads through data layout transformations and operator fusion, doing so is more challenging in a general DL framework, and we leave such optimizations to future work.
Nevertheless, \sten{} is faster than PyTorch, while retaining the productivity of a full DL framework.
Indeed, \sten{} enabled the rapid prototyping and testing of the \nofmg{} format, which could subsequently be integrated into inference engines to benefit from their optimizations.

\subsection{\sten{} Productivity}

While \cref{sec:sten} has shown that \sten{}'s interface is intuitive, we now demonstrate this practically by implementing several existing sparsifiers.
As our goal is to showcase productivity, we conduct a simple evaluation in which we fine-tune a model to 50\% sparsity using unstructured one-shot, iterative, and layer-wise magnitude pruning.
We consider the programming effort to implement the sparsifiers and associated sparse training.
We emphasize that we do not aim to outperform state-of-the-art pruning methods in this study.

\begin{figure}[t]
  \centering
  \includegraphics[width=\linewidth]{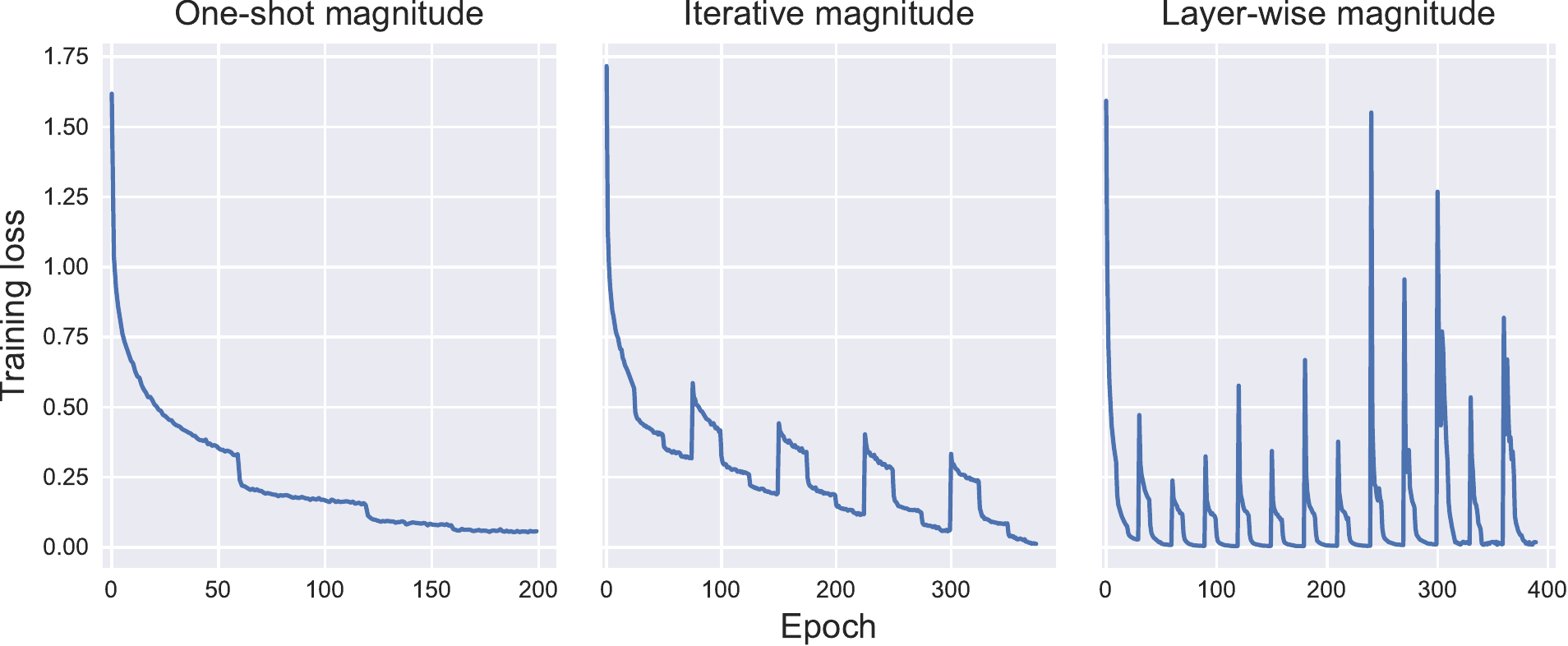}
  \caption{Training loss across sparsifiers for pruning a pretrained Wide ResNet-16-8 on CIFAR10 to 50\% sparsity.}\label{fig:resnet-pruning}
\end{figure}

We use a Wide ResNet-16-8~\cite{zagoruyko2016wide} trained on CIFAR10~\cite{cifar}, which consists of 11.0 M parameters.
Our dense model is trained following \citeauthor{zagoruyko2016wide}~\cite{zagoruyko2016wide}, with batch size 512 for 200 epochs on four 16 GB V100 GPUs.
We apply one-shot pruning and then fine-tune following the same schedule.
Iterative pruning begins at 10\% sparsity, then iteratively fine-tunes for 75 epochs before increasing the sparsity by 10\%.
Finally, layer-wise pruning prunes each layer and fine-tunes for 30 epochs, starting from the first layer.
We use one V100 GPU for fine-tuning, with batch size 128.
\Cref{fig:resnet-pruning} shows training curves for this process and \cref{tab:productivity} reports the final test accuracies.
Each method is able to approximately recover the dense accuracy, although one-shot pruning performs slightly worse.
Sparse training iterations are about 18\% slower than a dense iteration (median of 0.274 s verus 0.233 s per batch), showing the overhead of sparsity is low.

\begin{table}[t]
  \centering%
  \caption{Sparsification test accuracy and lines of code added.}\label{tab:productivity}%
  \footnotesize%
  \begin{tabular}{@{}lrr@{}}%
    \toprule
    Sparsifier & Top-1 Accuracy (\%) & LoC Added \\
    \midrule
    Dense & 95.02 & --- \\
    Sparsification setup & --- & 112 \\
    \quad One-shot magnitude & 94.84 & 6 \\
    \quad Iterative magnitude & 95.14 & 9 \\
    \quad Layer-wise magnitude & 95.10 & 9 \\
    \bottomrule
  \end{tabular}\vspace{0.5em}%
\end{table}

Implementing this sparsification is straightforward with \sten{}, and required no modification to the existing dense training loop, which is simply used with altered training schedules.
\Cref{tab:productivity} shows the number of lines of code added to implement sparse fine-tuning.
The sparsification setup consists of general components, including the implementation of magnitude pruning and a masked tensor which stores the pruning mask.
Sparse models are constructed using \sten{}'s \texttt{SparsityBuilder} (\cref{subsec:model-cons}).
This handles initial sparsification and storage, converting tensor types, and ensures that weights are re-sparsified after gradient updates.
Given this infrastructure, one-shot, iterative, and layer-wise magnitude pruning are then implemented using different different training and re-sparsification schedules, which requires only a few additional lines of code.

More generally, \sten{} supports sparsifying a broad range of models out-of-the-box, including torchvision's~\cite{torchvision} entire suite of image classification models, covering twenty model families and over a hundred pretrained models.  %
Adding support for sparse fine-tuning to torchvision's training code required only 25 lines of code.

\section{Related Work}
\label{sec:related}
There has been extensive work on sparsity in deep learning~\cite{hoefler2021sparsity}, and in scientific and high-performance computing in general~\cite{anzt2020preparing}.
Here we focus on key related work for deep learning sparsity systems.

\textbf{Systems for sparsity.}
General deep learning frameworks, such as TensorFlow~\cite{tensorflow2015-whitepaper} and PyTorch~\cite{paszke2019pytorch} provide basic support for sparse tensors and operations; more specialized frameworks, such as TVM~\cite{chen2018tvm,octomlsparse}, TFLite~\cite{tflitesparse}, and DeepSparse~\cite{kurtz2020inducing} aim to accelerate sparse inference.
Many works target sparsity for specific kernels or networks, rather than providing a complete pipeline~\cite{gale2020sparse,openaiblocksparse,pytorchblocksparse,park2016faster,elsen2020fast,yu2017scalpel,hill2017deftnn,kundu2020pre,lym2019prunetrain,li2017enabling,yao2019balanced,gong2020sparsetrain,chen2018escoin}; these could be incorporated into \sten{} for further performance improvements.
Another branch of work focuses on hardware acceleration for sparsity~\cite{dave2021hardware}.
Graph neural network systems also extensively leverage sparsity (e.g.,~\cite{besta2022parallel,hu2020featgraph,huang2020ge})

\textbf{Other model compression and acceleration approaches.}
There are many other approaches to model compression beyond sparsity; these include quantization~\cite{gholami2021survey}, parameter sharing (e.g.,~\cite{savarese2019learning,lan2019albert,plummer2020neural}), distillation~\cite{gou2021knowledge}, and factorization (e.g.,~\cite{wu2018prodsumnet,phan2020stable}).
Such methods are typically orthogonal and can be applied with sparsity for further improvements (e.g.,~\cite{li2022efficient,chen2021efficient}).

\section{Discussion}
\label{sec:discussion}
We introduced \sten{}, an interface for productive and efficient sparsity in PyTorch.
\sten{} is highly extensible and customizable, making it easy to explore sparsity.
Adding sparsity layouts or sparsifiers is simple, as is providing custom, optimized implementations for sparse operators.
Further, our new \nofmg{} sparsity layout provides performance competitive with highly-optimized sparse inference frameworks.
With \sten{}, sparsity ``just works'', typically taking only a handful of lines of code and allowing existing training code to be reused as-is.
This combination of features makes \sten{} an engine to help drive forward sparsity research and development within the broader machine learning community, allowing users to tackle problems in sparsity that previously would have been out of reach.

There are many potential future directions for study.
In particular, developing fully sparse training methods, where dense tensors are never materialized, is a major open problem~\cite{hoefler2021sparsity} that \sten{} allows researchers to make progress on.
We also see developing improved systems for sparsity as a major direction, especially for sparse training, where automatic optimizations (e.g., fusing sparsifiers) are less explored and potentially very valuable.

\section*{Acknowledgments}
This work has received funding from the European High Performance Computing Joint Undertaking (JU) under grant agreements No. 955513 (MAELSTROM) and No. 101034126 (EU-Pilot).
T.B.N. was supported by the Swiss National Science Foundation (Ambizione Project \#185778).
The experiments were conducted utilizing the resources of the Swiss National Supercomputing Centre.

\bibliographystyle{ACM-Reference-Format}
\bibliography{refs}

\end{document}